\documentclass[conference]{IEEEtran}
\usepackage{times}

\usepackage[numbers]{natbib}
\usepackage{multicol}
\usepackage[bookmarks=true]{hyperref}
\usepackage{graphicx}

\let\labelindent\relax 
\usepackage{enumitem}

\usepackage{amsmath}
\usepackage{xcolor}

\begin{document}

\title{Robot Policy Evaluation for Sim-to-Real Transfer: A Benchmarking Perspective}

\author{Xuning Yang, Clemens Eppner, Jonathan Tremblay, Dieter Fox, Stan Birchfield, Fabio Ramos\\ NVIDIA}

\maketitle

\begin{abstract}
Current vision-based robotics simulation benchmarks have significantly advanced robotic manipulation research. However, robotics is fundamentally a real-world problem, and evaluation for real-world applications has lagged behind in evaluating generalist policies. 
In this paper, we discuss challenges and desiderata in designing benchmarks for generalist robotic manipulation policies for the goal of sim-to-real policy transfer. We propose 1) utilizing high visual-fidelity simulation for improved sim-to-real transfer, 2) evaluating policies by systematically increasing task complexity and scenario perturbation to assess robustness, and 3) quantifying performance alignment between real-world performance and its simulation counterparts. 

\end{abstract}

\IEEEpeerreviewmaketitle

\section{Introduction}
Standardized evaluation has been crucial in the advancements of Large Language Models (LLMs) and Visual Language Models (VLMs).
Strategic benchmarks such as Massive Multitask Language Understanding (MMLU) \cite{hendrycks2021measuring} and Holistic Evaluation of Language Models (HELM) \cite{liang2022holistic} have presented a systematic way to represent language-based scenarios and evaluate trained policies against a set of diverse subjects.
In addition, they include soft-metrics such as robustness, fairness, and bias to help understand the performance beyond just successful responses. 
These efforts lead to the development of the useful language AI applications that we use today. 


In contrast, robotic evaluation for generalist manipulation policies has lagged behind, particularly for real-world applications. 
Current robotic benchmarks are characterized by specialized task suites with narrow focus, such as multi-task reinforcement learning \cite{rlbench, metaworld}, VLM-based robotic reasoning \cite{zhao2025manipbench}, and limited testing tasks \cite{libero}.
Moreover, most benchmarks lack considerations for robustness in deploying robot policies in the real world, which have been shown to significantly degrade policy performance~\cite{colosseum}.

\begin{figure}
    \centering
    \includegraphics[width=1\linewidth]{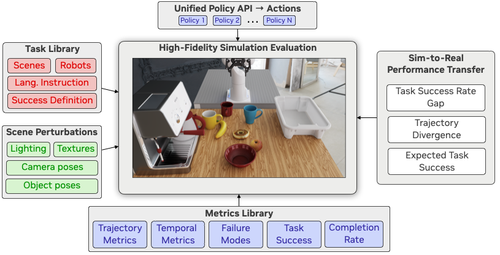}
    \caption{Overview of the proposed evaluation benchmark.}
    \label{fig:system}
    \vspace{-16pt}
\end{figure}

The absence of a standardized, scalable robotic benchmark for sim-to-real transferability presents a critical bottleneck for visual policy for robotics. In this paper, we discuss the key desiderata for a robotic benchmark aimed at training generalist robot policies, ensuring that real-world challenges, such as robustness and task difficulty, are effectively represented. 
We describe several discrete and continuous metrics, as well as potential tools to be used for comparing simulation benchmarks to real robot performance. 
Lastly, we outline our approach for a scalable benchmark system using high-fidelity simulation for systematically evaluating robotic policies (Fig.~\ref{fig:system}). 
We hypothesize that systematic simulation has the potential to enable \emph{scalable robotics benchmarking} as a viable proxy to extensive real-world experiments.

\section{Challenges in Robotics Benchmarking}
 \begin{figure*}
    \centering
    \includegraphics[width=1\linewidth]{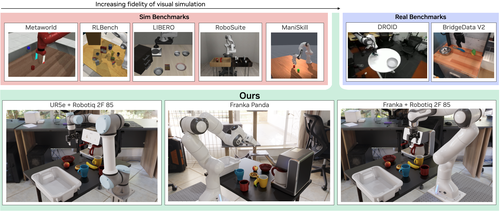}
    \caption{Comparison of visual fidelity across various simulated benchmarks and real-world datasets.}
    \label{fig:fidelity}
    \vspace{-12pt}
\end{figure*}





\subsection{Sim-to-Real} 
The sim-to-real gap remains a top challenge for vision-based policies. 
Transferring policies learned in simulation to real-world often fails due to various discrepancies in contact physics, visual appearance, and environmental dynamics with performance drop as high as 24--30\% \cite{simpler}. 
A common approach to both the visual and physical gap is to perform domain randomization \cite{libero, robosuite}.
Another approach is to combine synthetic and real data, which requires fine-tuning on a set of environment-specific data in order to increase the performance of the model \cite{pmlr-v155-julian21a}.

\textbf{Visual fidelity.} Visual fidelity plays a particularly critical role in this transfer. Traditional simulators often produce unrealistic visual observations that fail to capture the complexity of real-world lighting, textures, and environmental variations, as shown in Fig.~\ref{fig:fidelity}. When trained with lower-quality simulation images, policies deployed in the real world face significant performance drops ~\cite{khanna2023hssd,simpler}. However, with high-quality images, it is possible for the policy to transfer to the real world without any additional fine-tuning \cite{khanna2023hssd, meng2024amr}. This suggests that the level of photorealism in benchmarking environments is equally crucial for accurately evaluating sim-to-real policies. 

\textbf{Scene variation.}
Current datasets do not provide a systemic set of scene variations. However, recent works \cite{colosseum} have shown that changes in lighting and camera poses causes model success rate to degrade between 30--50\%. 


\subsection{Language Annotated Tasks} 
\begin{figure}
    \centering
    \includegraphics[width=0.8\linewidth]{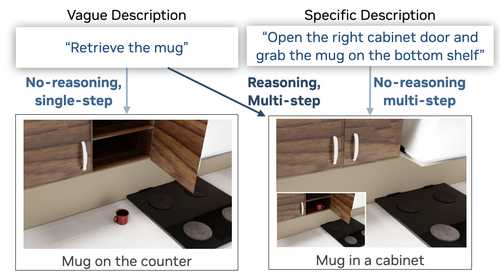}
    \caption{A simple command such as {\em ``retrieve the mug''} requires different levels of reasoning based on the scene. If the mug is in front of the robot, the task is fairly simple; however, if the mug is in a closed cabinet in a large kitchen, this would require open-world reasoning, including inferring the possible locations of the mug and recognizing and interacting with other objects in the scene such as cabinet doors in order to explore.}
    \label{fig:language}
    \vspace{-12pt}
\end{figure}

Robotics data has traditionally not focused on open-vocabulary instruction for tasks.
However, recent robotic models have leveraged VLMs for generalizing to specific robotic open-world task settings \cite{geminiroboticsteam2025geminiroboticsbringingai}. 
While some datasets incorporate natural language instructions \cite{rtx_2023, rlbench, bridge}, these datasets lack structured language annotations. 
As VLMs become prevalent in robotics, the language prompt itself may be a point of interest in future benchmarks, focusing on how well robots interpret and execute instructions that have varying levels of specificity~\cite{progprompt}. 
Complexity of the task changes as a function of the scene and language instruction. Depending on the specificity of the instruction, the task may or may not require open-world reasoning, as illustrated in Fig.~\ref{fig:language}.

\subsection{Unified Platform} 
Existing datasets follow lax task definitions which are inconsistent across datasets, which vary widely and create difficulty in cross-platform and cross-embodiment comparisons.  
Additionally, robotic policies output various action spaces (\textit{e.g.}, joint position/velocity, end effector pose \cite{robosuite}, or action primitives \cite{progprompt, codeaspolicies2022}), further complicating a standardized evaluation.
To develop a unified benchmarking platform, it needs to be representative of the real-world conditions: the task structure needs to be systematic in categorizing tasks based on complexity and skill, and the policy interface needs to support multiple types of actions in order to  enable a consistent comparison of policy architectures on identical tasks.

\subsection{Scale and Scope}
Unlike datasets used to train LLM and VLMs, robotics data is difficult to obtain. Recent open-sourced initiatives such as RT-X \cite{rtx_2023} and DROID \cite{droid} have focused on using teleoperation to collect large-scale data in the real-world. For simulated data, aggregation frameworks such as RoboVerse \cite{geng2025roboverse} aim to unify benchmarks. 

While large-scale datasets are useful in training, the suite of benchmarking tasks is not necessarily large in size but comprehensive in scope. Effective sim-to-real benchmarks need to have a curated set of tasks and environments that systematically cover a broad spectrum of skills, perceptual challenges, and task complexities. 
\citet{khanna2023hssd} show that a small set of high-quality scenes ($\approx 200$) can outperform larger procedurally generated scenes ($\approx 10k$) in policy learning.

Thus, \emph{scale} in benchmarking tasks needs to focus on diversity and real-world relevance. 
High-fidelity simulation enables systematic variations in representative tasks, which enables precise, repeatable benchmarking across a wide range of realistic scenarios. As real-robot data collection becomes less scalable as the field progresses, simulation offers a sustainable path forward.



\section{Desiderata for Robot Manipulation Benchmarking} \label{sec:desiderata}
Given the challenges above and the segregated landscape of existing frameworks, 
we present a recipe for a benchmarking framework for evaluating vision-based robotics policies. We employ the following definitions: A \emph{task} $\mathcal{T}$ is a set of motions or \emph{sub-tasks}, $\tau$, that completes a \emph{language-based instruction},~$l$ \cite{bridge}. We consider single-manipulator robot tasks, with the policy $\pi\!:\!\mathcal{O}\!\rightarrow\!\mathcal{A}$ where the action space $\mathcal{A}$ is policy dependent.

\subsection{Task Taxonomy} \label{sec:desiderata:taxonomy}

%
We introduce a novel task taxonomy that systematically categorizes tasks based on increasing complexity, required skills, and generalization. 
We categorize tasks into four difficulty levels:

\begin{enumerate}[label=\textbf{T\arabic*},start=1]
    \item \textbf{Single-motion tasks:} (e.g., pick, place, open, close) These typically involve a single, well-constrained action primitive involving a visually present object. These tasks test core visual reasoning and visuomotor capabilities. In particular, \emph{pick/place} require the robot to reason about stable grasps; and \emph{open/close} require reasoning about the joint constraints of the fixture (e.g., door hinges, sliders).
    
    \item \textbf{Continuous-motion tasks:} (e.g., wiping, stirring, or pouring) These require smooth trajectories and precise control over a constrained space. These tasks require the robot to reason about tool-use and the space in which the continuous motion is constrained within. 
 
    \item \textbf{Multi-step tasks:} (e.g., put away, clean up) These combine multiple primitives into a temporally extended sequence of skills, which often require open-world reasoning of the scene and planning under partial observability and long-horizon dependency.

    \item \textbf{Long-horizon tasks with memory:} Lastly, we consider cases where the robot needs to reason about its broader environment over its global memory\footnote{This would be akin to Retrieval-Augmented Generation (RAG) \cite{lewis2021ragnlp} mechanisms employed in LLMs.}. These type of tasks require the robot to retain memory of objects' spatial relationships over time. These type of tasks are typical of mobile manipulators, where a task may involve retrieving objects from multiple locations. 
\end{enumerate}

\subsection{Robustness to Variations}
To evaluate the robustness of a policy, it is important to apply a range of systematic perturbations to the environments. 
We introduce a suite of variations (Fig.~\ref{fig:variations}) to simulate diverse deployment conditions, which may emerge during deployment in dynamically changing environments, following \cite{simpler}:

\begin{figure*}
    \centering
    \includegraphics[width=1\linewidth]{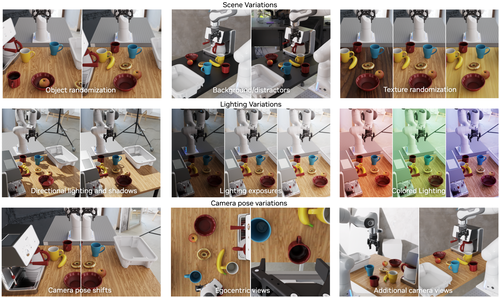}
    \caption{Example scene variations, lighting variations, and camera pose variations. These are common perturbations present in real world settings.}
    \label{fig:variations}
    \vspace{-16pt}
\end{figure*}

\begin{enumerate}[label=\textbf{V\arabic*},start=1]
    \item \textbf{Object placement:} Object positions are perturbed within the workspace following a pre-specified distribution. These shifts assess the policy's ability to handle spatial displacements. 
    \item \textbf{Number of Objects:} This introduces distractor and occlusions that test the model's ability to distinguish relevant objects. 
    \item \textbf{Texture changes:} Surface textures of objects and background are randomized using a library of synthetic and real-world backdrops. These variations assess the policy’s reliance on appearance-specific features and its ability to generalize across visually distinct but semantically identical environments.
    \item \textbf{Lighting changes:} Altering conditions (e.g., ambient light intensity, directional light and shadow) challenges the visual encoder’s robustness to changes in illumination. 
    \item \textbf{Camera pose variations:} 
    Even small discrepancies in camera pose between training and deployment can lead to significant performance degradation, making robustness to pose variation critical for practical reliability. Therefore, perturbations in camera viewpoints tests the stability of the policy under pose deviations.
\end{enumerate}


\subsection{Discrete and Continuous Metrics} \label{sec:desiderata:metrics}
Evaluations in robot policy learning have traditionally focused on task success rates, but this binary metric often fails to capture the full spectrum of policy performance~\cite{kressgazit2024bestpractices}. 
To address this limitation, we define a more granular set of metrics including discrete and continuous metrics. These metrics allow us to understand policy behavior and limitations in end-to-end robot learning.

\begin{enumerate}[label=\textbf{M\arabic*},start=1]
    \item \textbf{Completion Rate $\mathcal{C}$:} The percentage of successful task completions in a total set of attempted tasks $\mathcal{C}(\pi)$. This quantifies the ability for the policy to complete the task from start to finish.
    \item \textbf{Task Success $\mathcal{S}$:} We reframe this to describe the percentage of \emph{sub-tasks} that have been completed: $\mathcal{S}(\mathcal{T})=\frac{1}{\mathcal{T}} \sum_{\tau \in \mathcal{T}} \mathcal{S}(\tau)$
    This approach provides a graded measure of success.
    \item \textbf{Failure Modes:} These are systematically categorized to enable precise diagnosis of failure cases~\cite{agia2024unpackingfailuremodesgenerative, robofail}: 
    {\small
    \begin{enumerate}
    \item \emph{Grasp Failure:} The robot fails to establish initial contact, often due to inaccurate pose estimation, poor alignment, or insufficient gripper closure.
    \item \emph{Grasp Stability Failure (Object Dropped):} The robot successfully grasps the object but subsequently loses it due to an unstable grasp.
    \item \emph{Policy Generation Failure:} The policy outputs invalid and infeasible actions.
    \item \emph{Reachability Failure:} The target action is unreachable due to the robot’s kinematic constraints. 
    \item \emph{Reasoning Failure:} The robot exhibits incorrect high-level decision-making or planning, such as selecting inappropriate actions or misinterpreting task goals. 
    \end{enumerate}
    }


    \item \textbf{Trajectory Metrics. } Trajectory metrics capture quality, efficiency, and desirability of robot motion.
    {\small
    \begin{enumerate}
    \item \emph{Path Length:} The total distance traveled by the robot’s end-effector during task execution. 
    \item \emph{Trajectory Smoothness:} Quantifies the consistency and fluidity of motion, measured by the higher derivatives of the trajectory.
    \item \emph{Trajectory Optimality:} Quantifies whether the actions were time optimal; or if any corrective actions were taken. 
    \end{enumerate}
        }
    \item \textbf{Temporal Metrics. } Temporal metrics capture time efficiency of task execution:
    {\small
    \begin{enumerate}
    \item \emph{Total Time to Completion:} This measures system throughput and operational speed.
    \item \emph{Average Policy Inference Time:} The measures the ability for the policy to be deployed online.
    \item \emph{Episode Duration:} The total time span of an entire task attempt, including all actions and any recovery or correction phases.
    \end{enumerate}
    }

\end{enumerate}

\subsection{Sim-to-Real Transfer}
We introduce quantitative metrics to evaluate the \emph{transfer fidelity} of a policy's performance in sim vs. real.
\begin{enumerate}[label=\textbf{S\arabic*},start=1]
    \item \textbf{Success Performance Matching in Fixed Baselines.} This refers to the difference in success rates between simulation and real-world deployment for a set of controlled scene equivalents. For a single task, this can be measured by $\|\mathcal{S}_\text{sim}(\pi)-\mathcal{S}_\text{real}(\pi)\|^2$, where $\|\cdot\|^2$ is the $L^2$ norm. However, for a range of tasks, Mean Maximum Rank Violation (MMRV) \cite{simpler} has been used to describe performance shifts due to scene variations.
    \item \textbf{Trajectory Performance Matching.} We utilize \emph{trajectory divergence}, a metric defined over the state evolution between trajectories executed in simulation and in real. For each motion $\tau=({s_0,s_1,\dots,s_T})$, the divergence is given by $D(\{\tau_\text{sim}^i\}_{i=1}^N,\{\tau_\text{real}^j\}_{j=1}^M)$. Potential choices for $D(\cdot,\cdot)$ include the Maximum Mean Discrepancy (MMD)~\cite{gretton2012kernel}, energy statistics~\cite{szekely2013energy}, and the classical Friedman-Rafsky test~\cite{Friedman1979Multivariate}. 
    \item \textbf{Expected Success Rate in Real.} Given the success rate obtained in simulation, it is possible to estimate the probability that the success rate for a policy $\pi$ in real is higher than a threshold $\theta$, as the posterior
    $p(\mathcal{S}_\text{real}(\pi)>\theta|\mathcal{S}_\text{sim}(\pi), \mathcal{T}, l)\propto p(\mathcal{S}_\text{sim}(\pi)|\mathcal{S}_\text{real}(\pi)>\theta, \mathcal{T}, l)p(\mathcal{S}_\text{real}(\pi))$, if the simulator provides ground-truth values. We aim to quantify this using our future experiments.
\end{enumerate}

\section{Proposed Benchmarking Framework}
We discuss our initial efforts towards developing a vision-based robotic benchmarking framework, aimed at systematically evaluating robotic policies for improving sim-to-real transfer performance. The core objective of this framework is to 1)~establish standardized protocols and metrics that evaluate vision-based policies in scalable high-fidelity simulation environments; and 2) quantify performance alignment between real-world experiments with its simulated equivalents, as described in Fig.~\ref{fig:system}. 

We propose leveraging a high-fidelity visual simulator (IsaacLab) to bridge the visual perception gap between simulation and real-world. We procedurally generate tasks according to Sec.~\ref{sec:desiderata:taxonomy}, including scenes, language descriptions, task-success definitions. Additionally, the benchmark contains a suite of scene perturbations addendums, used to randomize the task library.
and a suite of metrics as described in Sec.~\ref{sec:desiderata:metrics}. 

Using our proposed benchmarking framework, we plan to perform a set of real-world experiments complementary to sim and a comprehensive analysis of the performance gap. By comparing sim-to-real performances using proposed metrics, our framework facilitates the identification of specific failure modes and evaluation domain gap.
With this framework, we aim to increase systematic evaluation of robotic policies that scales as the field evolves. 
Ultimately, this framework is intended to serve as a reference pipeline for the broader community working on sim-to-real robot learning, including cross-comparison and reproducibility.

\bibliographystyle{plainnat}
\bibliography{references}

\end{document}